\documentclass[conference]{IEEEtran}
\IEEEoverridecommandlockouts
\usepackage{cite}
\usepackage{tikz}
\usepackage{seqsplit}
\usetikzlibrary{arrows.meta, positioning, calc}
\usepackage{algorithm}
\usepackage{algorithmic}
\usepackage{subcaption}
\usepackage{amsmath,amssymb,amsfonts}
\usepackage{colortbl}
\usepackage{comment}
\usepackage{textcomp}
\usepackage{stfloats}
\usepackage{graphicx}
\usepackage{booktabs}
\usepackage{hyperref} 
\usepackage{multirow} 
\usepackage[table]{xcolor}

\tiny
\def\BibTeX{{\rm B\kern-.05em{\sc i\kern-.025em b}\kern-.08em
    T\kern-.1667em\lower.7ex\hbox{E}\kern-.125emX}}

\newcommand{\name}{DeFiRank}
\begin{document}

\title{\name: Federated Pairwise Ranking for Governance in Decentralized Financial Networks

}

\author{
    \IEEEauthorblockN{
        Zahra Batool\IEEEauthorrefmark{1}\IEEEauthorrefmark{3}, 
        Michael Duchesne\IEEEauthorrefmark{2}, 
        Kaiwen Zhang\IEEEauthorrefmark{1},
        Ulrich Aivodji\IEEEauthorrefmark{1}\IEEEauthorrefmark{3}
    \IEEEauthorblockA{\IEEEauthorrefmark{1}École de technologie supérieure, Montréal, Canada}
    \IEEEauthorblockA{\IEEEauthorrefmark{2}T-RIZE GROUP, Montréal, Canada}
    \IEEEauthorblockA{\IEEEauthorrefmark{3}Mila - Quebec AI Institute, Montréal, Canada}\\
Email: \{zahra.batool, kaiwen.zhang, Ulrich.aivodji\}@etsmtl.ca, michael.duchesne@t-rize.io}
}

\maketitle

\begin{abstract}
Existing governance approaches in decentralized financial networks rely on manual review processes conducted by human committees, which introduces subjective bias, creates a scalability bottleneck under high network volume, and fails to capture complex behavioural patterns and evolving network dynamics. To address these limitations, we propose \name, a federated learning-based pairwise ranking framework in which multiple institutions collaboratively train a ranking model without sharing their private transaction data. This collaborative learning enables the model to capture network-wide behavioural patterns that signify strong network participation and directly contribute to the network effect, including diversity of counterparties, usage of multiple applications, and frequency of transactions. The proposed system ranks providers based on the authenticity of their actions, promoting genuine contributors and demoting those exhibiting suspicious behaviour. We evaluate \name\ on a real-world blockchain transaction dataset. Results demonstrate that the federated approach achieves ranking performance modestly below the centralized baseline (NDCG@15: $\approx$0.87 vs.\ $\approx$0.92) while preserving data privacy and matching its ranking stability (Kendall $\tau$: $\approx$0.70 for both regimes), producing reliable governance decisions, including provider promotion and demotion (648 and 39 unique providers, respectively), without requiring centralized access to sensitive transaction data.

\end{abstract}


\section{Introduction} 
Corporate governance in decentralized financial networks refers to the set of rules and mechanisms through which entities are regulated and overseen by financial institutions and intermediaries to maintain transparency, stability, and accountability \cite{munir2025ai}. Traditionally, corporate governance has relied heavily on manual processes and human judgment, making it slow, inconsistent, and difficult to scale across large decentralized networks \cite{aguilera2025toward}. Machine learning (ML) has since transformed governance, enabling automated, data-driven decisions at scale \cite{ma2025algorithmic}.

Traditionally, training an ML model requires data to be collected at a central server \cite{batool2025privacy}. 
Data privacy regulations such as GDPR restrict how sensitive financial data may be gathered and shared, creating tension with centralized training: dispersed financial institutions must relinquish ownership of proprietary transaction records to a single server, raising data sovereignty and institutional trust concerns.

The underlying problem is fundamentally one of reputation and Sybil-resistance: determining whether a given provider is a genuine, authentic contributor to the network, as opposed to a manipulative or low-value actor. Approaches to this problem in the literature include automated scoring mechanisms analogous to Elo or Glicko rating systems \cite{glickman1995glicko, berg2020statistical}, as well as Sybil-tolerant reputation systems designed specifically for decentralized governance and tokenomics \cite{nasrulin2022meritrank}. In practice, however, current decentralized financial networks address this problem through manual governance: human committees periodically evaluate providers, promoting authentic contributors to featured status and demoting others \cite{polcumpally2024blockchain}. This provides stable rankings but depends on a human-crafted, single-party scoring function with predetermined weights, introducing systemic bias and struggling to adapt as network behaviour and applications evolve, creating a lag between network dynamics and governance decisions.


Crucially, most provider-to-provider interactions in a decentralized financial network are private and never observed by any single centralized entity. As a result, the network-wide patterns indicative of authentic behaviour, such as the global distribution of application usage or typical counterparty diversity, cannot be learned from a central vantage point. Federated Learning (FL) is therefore a natural fit for this setting: it allows a global ranking model to learn from patterns emerging across private, localized transactions held by multiple hosts, without ever requiring those hosts to expose their raw data \cite{mcmahan2017communication, aledhari2020federated}. This preserves institutional privacy while enabling the capture of network-wide behavioural trends that are invisible to any single actor. 

In this context, authenticity of actions refers to how genuine and legitimate a provider's behaviour is within the network. The objective is to rank providers such that those engaging in diverse, normal financial activity score higher, while providers exhibiting suspicious or highly constrained behaviour score lower.

Learning-to-rank is a well-established paradigm in information retrieval, with pairwise methods receiving particular attention for their ability to directly optimise relative ordering between entities \cite{burges2005learning, liu2009learning}. Pairwise ranking trains a model to minimise ranking inversions, that is, to reduce mistakes in the relative preference ordering between pairs \cite{liu2009learning}. In corporate governance, this translates naturally to comparative decisions: given two providers, which demonstrates higher authenticity? This formulation allows the model to identify which features, such as counterparty diversity, transaction volume, or application engagement, are most indicative of genuine network contribution. 

To overcome the limitations of existing manual scoring approaches, we propose \textbf{\name}, a federated pairwise \textbf{rank}ing framework for governance in \textbf{de}centralized \textbf{fi}nancial networks. \name\ replaces the manual epoch scoring function with a machine-learned ranker that captures the features signifying strong network participation, including diversity of counterparties, usage of multiple applications, and frequency of transactions, while preserving the privacy of each provider's transaction data. The main contributions of this work are: 

\begin{enumerate} 
\item We propose a privacy-preserving federated pairwise ranking framework that enables decentralized financial networks to collaboratively learn provider rankings without sharing private transaction data (Section~\ref{section:pm}). 
\item We design a dynamic governance decision pipeline with adaptive promotion and demotion thresholds that respond continuously to network health (quality trends) and network size (growth and contraction) (Section~\ref{subsec:Governance_decision_making}). 
\item We demonstrate through experiments that the federated approach closely matches centralized ranking stability (Kendall~$\tau$: $\approx$0.70 for both regimes) while trailing modestly in top-rank precision (NDCG@15: $\approx$0.87 vs.\ $\approx$0.92 centralized), and producing trustworthy governance decisions comprising 648 promotion and 39 demotion candidates, with 100\% promotion consistency and 72\% demotion consistency over 2+ consecutive days (Section~\ref{sec:results}).

\end{enumerate} 
The remainder of this paper is organised as follows. Section~\ref{section:LR} reviews related work. Section~\ref{section:pm} describes the proposed \name\ architecture. Section~\ref{section:pe} presents experimental results and comparisons with baseline approaches. Finally, Section~\ref{sec:conclusion} concludes the paper and discusses directions for future work.

\section{Literature Review}
\label{section:LR}
This section reviews previous related work on pairwise ranking in federated learning and AI-driven corporate governance.

\subsection{Pairwise Ranking in Federated Learning}
Federated learning-to-rank remains underexplored, particularly in financial and governance contexts. \cite{wang2021effective} introduces the Federated Pairwise Differentiable Gradient Descent (FPDGD) protocol for online learning-to-rank. 
\cite{anelli2021federated} applies federated pairwise ranking to recommendation systems, proposing a protocol based on matrix factorization and Bayesian personalized ranking. \cite{hartmann2019federated} implements point-wise federated ranking to optimize Firefox address-bar URL suggestions, keeping browsing history and input on-device throughout training. \cite{guo2023fedxl} addresses compositional pairwise risk optimization in federated learning through gradient decoupling. These methods target query-document retrieval or recommendation, where labels derive from clicks or ratings. 

\subsection{AI-Driven Corporate Governance}
AI-driven governance has gained traction for reducing human bias and improving decision-making consistency. \cite{mahajan2025artificial} shows AI-based frameworks improve real-time decisions across complex business ecosystems, reflecting a broader shift from rule-based to learning-based governance \cite{munir2025ai}. However, existing work predominantly assumes centralized data access, limiting applicability to decentralized settings. \name\ addresses this gap by combining learned governance scoring with the privacy guarantees of federated learning.

The aforementioned studies provide the algorithmic machinery for privacy-preserving ranking but have not been applied to governance settings where heterogeneity is structural and outputs drive real promotion and demotion decisions, while the AI-governance literature addresses governance objectives but assumes centralized data access. \name\ is positioned at the intersection of these two gaps: a federated pairwise ranking framework purpose-built for decentralized financial governance.

\section{Proposed Method}
\label{section:pm}
\name\ involves three entities: 
\begin{enumerate}
   
 \item  \textbf{\textit{Host:}} a data-holding institution, used interchangeably with federated client.
 \item \textbf{\textit{Provider:}} the governed unit ranked and promoted/demoted, whose daily features are computed at a single host and never combined across hosts.
 \item \textbf{\textit{Party:}} end user, whose activity aggregates into provider features. 
 \end{enumerate}
 \name\ operates only on provider-day aggregated features (Section~\ref{subsec:gov_score}) and never observes party-level data or data from another host. It consists of three phases: (1) \textit{federated temporal pairwise ranking}, partitioning the provider dataset across non-IID clients with structurally distinct governance perspectives; (2) \textit{local training}, where each client trains a temporal ranking model and the server aggregates updates via FedAvg (Section~\ref{subsec:local_training}); and (3) \textit{governance decision making}, where the trained model's scores drive data-driven promotion and demotion decisions (Section~\ref{subsec:Governance_decision_making}). Fig.~\ref{fig:overview} provides an overview of the full pipeline.

\begin{figure}[h]
    \centering
    \resizebox{\columnwidth}{!}{
\definecolor{cPink}{HTML}{FBD0CE}
\definecolor{cPinkBorder}{HTML}{E08A86}
\definecolor{cOrange}{HTML}{F8D49B}
\definecolor{cOrangeBorder}{HTML}{D99B3E}
\definecolor{cTeal}{HTML}{9ECAC9}
\definecolor{cTealBorder}{HTML}{4F8C8A}
\definecolor{cGreen}{HTML}{AEE0AE}
\definecolor{cGreenBorder}{HTML}{5BA85B}
\definecolor{cPurple}{HTML}{D6C6E1}
\definecolor{cPurpleBorder}{HTML}{9B7EB0}

\begin{tikzpicture}[
    box/.style={
        rectangle,
        rounded corners=3pt,
        draw,
        thick,
        text=black,
        font=\sffamily\bfseries\scriptsize,
        align=center,
        minimum width=2.0cm,
        minimum height=0.8cm,
    },
    serverbox/.style={
        rectangle,
        rounded corners=3pt,
        draw,
        thick,
        text=black,
        font=\sffamily\bfseries\small,
        align=center,
        minimum width=10.6cm,
        minimum height=0.8cm,
    },
    arrow/.style={-{Triangle[length=2.5mm,width=1.8mm]}, thick, black},
    node distance=0.6cm and 0.35cm
]

\node[box, fill=cPink, draw=cPinkBorder] (c1) {Client 1\\Reciprocity};
\node[box, fill=cPink, draw=cPinkBorder, right=of c1] (c2) {Client 2\\Engagement};
\node[box, fill=cPink, draw=cPinkBorder, right=of c2] (c3) {Client 3\\Churn};
\node[box, fill=cPink, draw=cPinkBorder, right=of c3] (c4) {Client 4\\Volume};
\node[box, fill=cPink, draw=cPinkBorder, right=of c4] (c5) {Client 5\\Interconnectivity};

\node[box, fill=cOrange, draw=cOrangeBorder, below=of c1] (t1) {Local\\TemporalRankNet};
\node[box, fill=cOrange, draw=cOrangeBorder, below=of c2] (t2) {Local\\TemporalRankNet};
\node[box, fill=cOrange, draw=cOrangeBorder, below=of c3] (t3) {Local\\TemporalRankNet};
\node[box, fill=cOrange, draw=cOrangeBorder, below=of c4] (t4) {Local\\TemporalRankNet};
\node[box, fill=cOrange, draw=cOrangeBorder, below=of c5] (t5) {Local\\TemporalRankNet};

\node[serverbox, fill=cTeal, draw=cTealBorder, below=1.1cm of t3] (server)
    {Server: FedAvg Aggregation $\rightarrow$ Global Model};

\node[box, fill=cGreen, draw=cGreenBorder, below=1.0cm of server, xshift=-4.2cm] (scores)
    {Daily governance\\scores};
\node[box, fill=cGreen, draw=cGreenBorder, right=of scores] (thresh)
    {Adaptive $\tau_t$, $K_t$\\(Algorithm~1)};
\node[box, fill=cPurple, draw=cPurpleBorder, right=of thresh] (promo)
    {Promotion list\\$\mathcal{P}$};
\node[box, fill=cPurple, draw=cPurpleBorder, right=of promo] (demo)
    {Demotion list\\$\mathcal{D}$};

\draw[arrow] (c1) -- (t1);
\draw[arrow] (c2) -- (t2);
\draw[arrow] (c3) -- (t3);
\draw[arrow] (c4) -- (t4);
\draw[arrow] (c5) -- (t5);

\draw[arrow] (t1.south) -- ($(t1.south)+(0,-0.3)$) -| (server.north);
\draw[arrow] (t2.south) -- ($(t2.south)+(0,-0.3)$) -| (server.north);
\draw[arrow] (t3.south) -- (server.north);
\draw[arrow] (t4.south) -- ($(t4.south)+(0,-0.3)$) -| (server.north);
\draw[arrow] (t5.south) -- ($(t5.south)+(0,-0.3)$) -| (server.north);

\draw[-{Triangle[length=2.5mm,width=1.8mm]}, thick, cTealBorder, dashed]
    (server.west) .. controls +(-2.0,0.3) and +(-1.5,-0.3) .. (c1.west)
    node[midway, above, font=\sffamily\scriptsize, text=cTealBorder, xshift=-0.3cm] {broadcast weights};

\draw[arrow] (server.south) -- ++(0,-0.4) -| (scores.north);
\draw[arrow] (scores) -- (thresh);
\draw[arrow] (thresh) -- (promo);
\draw[arrow] (promo) -- (demo);

\end{tikzpicture}}
    \caption{Overview of \name} 
    \label{fig:overview}
\end{figure}

\subsection{Federated Temporal Pairwise Ranking}

To evaluate \name\ under realistic multi-institution conditions, we simulate a federated deployment by partitioning the provider dataset, comprising 965,894 provider-day records spanning 102,520 unique providers over 184 days, across 5 non-IID clients. We emphasize that this partitioning is a methodological step for experimentation only: in a real deployment, no such partitioning would be required, as each client would natively hold its own private transaction data and never observe the data held by other clients. Here, the non-IID structure is constructed to reflect realistic behavioral heterogeneity across institutions; each client is assigned a different validator weight vector that prioritizes a specific governance dimension: reciprocity-heavy, engagement-heavy, churn-heavy, volume-heavy, and interconnectivity-heavy (formalized in Eq.~\ref{eq:client_weights}). This means the same provider pair can receive opposite labels across clients, since each evaluates provider quality from a different structural perspective, approximating realistic conditions where institutions prioritize different behaviour patterns based on their own vantage point.

\subsection{Local Training}
\label{subsec:local_training}

Local training is performed for pairwise ranking using TemporalRankNet. For each provider, \textit{build\_sequences()} constructs a sliding 7-day rolling window of 13 governance features (10 base + 3 trend), yielding input sequences of shape $(N, 7, 13)$. Early days with insufficient history are zero-padded to maintain consistent sequence length. Pairwise training labels are generated by comparing governance scores of provider pairs within the same day using each client's structural weight vector, yielding a total of 9,657,042 training pairs across all clients.

Every client maintains a local TemporalRankNet model combining LSTM temporal encoding with RankNet pairwise ranking. A two-layer LSTM processes each provider's 7-day sequence day by day, and the final hidden state $h_7$ yields a 64-dimensional embedding of the provider's governance behavior over the window. RankNet takes two same-day providers $A$ and $B$, encodes each independently via the shared LSTM, and optimizes pairwise binary cross-entropy loss:

\begin{equation}
\mathcal{L} = \text{BCE}\!\left(s_A - s_B,\ y_{AB}\right)
\label{eq:pairwise_loss}
\end{equation}

\noindent where $y_{AB} = 1$ if provider $A$ has higher governance quality than $B$, and 0 otherwise.

Each communication round proceeds as follows: (1) the server broadcasts the current global model weights to all 5 clients, (2) each client trains locally on its partitioned data for one local epoch using the BCE pairwise loss with the Adam optimizer ($\eta = 0.001$, batch size 2048), (3) clients transmit updated weights back to the server, no raw data, labels, or gradients are shared, preserving data privacy, (4) the server aggregates all client weights via FedAvg, computing a weighted average proportional to each client's dataset size. After aggregation, server-side evaluation computes NDCG@5/10/15 and Kendall $\tau$ using a neutral validator scoring function with a fixed seed, deliberately independent of any individual client's local weight distribution. Training runs for 100 communication rounds, with the global model converging within the first 10 rounds.

\subsection{Governance Decision Making}
\label{subsec:Governance_decision_making}

After training, the model assigns a predicted governance score to each provider daily. Algorithm~\ref{alg:governance} summarizes the four-stage pipeline used to convert these scores into promotion and demotion decisions over a $B=30$-day baseline period.

\textbf{Baseline anchors.} The demotion anchor $K_{\text{base}}$ is set at the largest score gap among sorted baseline featured apps (line~\ref{line:demotion_anchor}). The promotion anchor $\tau_{\text{promo}}$ escalates from the 25th to 50th percentile of baseline active-featured scores, then to a ratio-based estimate, if pinned at the 0.10 floor. The minimum ratio anchor $\rho_{\text{min}}$ is the median promotion ratio among baseline qualifiers above $\max(\tau_{\text{promo}}, 0.30)$, raised further if $\tau_{\text{promo}}$ saturates at its 0.60 ceiling; this yields $\rho_{\text{min}}=0.6231$ in our dataset.

\textbf{Daily scaling.} Both anchors are scaled daily by the network health and size ratios (Algorithm~\ref{alg:governance}, lines~\ref{line:promo_threshold}, \ref{line:effective_topk}): the promotion threshold $\tau_t$ rises when the network is thriving and falls when it is struggling, while the demotion cutoff is scaled by network size and bounded by the current featured-set size.

\textbf{Promotion and demotion.} Featured apps with near-zero activity are excluded before computing promotion ratios, preventing inactive competitors from inflating non-featured apps' apparent quality. A non-featured app is a raw promotion candidate if its ratio meets $\tau_t$; candidates are confirmed only if they qualify on 2+ consecutive calendar days with mean ratio at or above $\rho_{\text{min}}$ (line~\ref{line:reliability}), filtering transient signals from sustained ones. Featured apps ranked below the effective cutoff are flagged for demotion.
\begin{algorithm}[t]
\small
\caption{Dynamic Governance Decision Pipeline}
\label{alg:governance}
\begin{algorithmic}[1]
\REQUIRE Daily provider scores, featured-status flags, activity; baseline period $B=30$ days
\ENSURE Promotion list $\mathcal{P}$, demotion list $\mathcal{D}$
\STATE \textit{Step 1: Baseline anchors (first $B$ days)}
\STATE $\tau_{\text{promo}} \gets$ 25th percentile of active featured scores
\IF{stuck at floor (0.10)} \STATE retry at 50th percentile \ENDIF
\IF{still stuck} \STATE fall back to 75th percentile of outrank ratios \ENDIF
\STATE Clip $\tau_{\text{promo}}$ to $[0.10, 0.60]$
\STATE $K_{\text{base}} \gets$ rank with the largest score gap between featured providers \label{line:demotion_anchor}
\STATE $\rho_{\text{min}} \gets$ median outrank ratio of providers clearing $\tau_{\text{promo}}$, clipped to $[0.25, 0.70]$
\IF{$\tau_{\text{promo}}$ stuck at ceiling (0.60)} \STATE raise $\rho_{\text{min}}$ to $\geq 0.685$ \ENDIF
\medskip
\STATE \textit{Step 2: For each day}
\FOR{each day}
    \STATE Scale $\tau_{\text{promo}}$ by today's health ratio $\to$ today's threshold, clipped to $[0.10,0.60]$ \label{line:promo_threshold}
    \STATE Scale $K_{\text{base}}$ by today's size ratio $\to$ today's cutoff, clipped to $[3,150]$, capped by featured count \label{line:effective_topk}
    \STATE Drop near-inactive featured providers; skip day if fewer than 3 remain
    \STATE Flag featured providers below the cutoff for $\mathcal{D}$
    \FOR{each non-featured provider}
        \STATE Flag as candidate if its outrank fraction meets today's threshold
    \ENDFOR
\ENDFOR
\medskip
\STATE \textit{Step 3: Reliability check}
\FOR{each candidate}
    \STATE Group qualifying days into runs of consecutive dates
    \IF{a run has $\geq 2$ days and average ratio $\geq \rho_{\text{min}}$} \label{line:reliability}
        \STATE Confirm provider for $\mathcal{P}$
    \ENDIF
\ENDFOR
\RETURN $\mathcal{P}, \mathcal{D}$
\end{algorithmic}
\end{algorithm}

\begin{figure}[h]
    \centering
    \resizebox{\columnwidth}{!}{

\definecolor{cPink}{HTML}{FBD0CE}
\definecolor{cPinkBorder}{HTML}{E08A86}
\definecolor{cOrange}{HTML}{F8D49B}
\definecolor{cOrangeBorder}{HTML}{D99B3E}
\definecolor{cTeal}{HTML}{9ECAC9}
\definecolor{cTealBorder}{HTML}{4F8C8A}
\definecolor{cGreen}{HTML}{AEE0AE}
\definecolor{cGreenBorder}{HTML}{5BA85B}

\begin{tikzpicture}[
    box/.style={
        rectangle,
        rounded corners=3pt,
        draw,
        thick,
        text=black,
        font=\sffamily\bfseries\small,
        align=center,
        minimum width=2.2cm,
        minimum height=0.8cm,
    },
    arrow/.style={-{Triangle[length=3mm,width=2mm]}, thick, black},
    node distance=1.0cm and 1.0cm
]

\node[box, fill=cPink, draw=cPinkBorder] (parquet) {Parquet files};
\node[box, fill=cPink, draw=cPinkBorder, right=of parquet] (csv) {Combined CSV};
\node[box, fill=cPink, draw=cPinkBorder, right=of csv] (clean) {Cleaning and\\encoding};

\node[box, fill=cOrange, draw=cOrangeBorder, below=of parquet] (trend) {Trend features\\(7-day)};
\node[box, fill=cOrange, draw=cOrangeBorder, right=of trend] (norm) {Feature\\normalization};
\node[box, fill=cOrange, draw=cOrangeBorder, right=of norm] (valid) {Validator\\weights};

\node[box, fill=cTeal, draw=cTealBorder, below=of trend] (gov) {Governance\\score};
\node[box, fill=cTeal, draw=cTealBorder, right=of gov] (seq) {7-day sequence\\windows};
\node[box, fill=cTeal, draw=cTealBorder, right=of seq] (pair) {Pairwise dataset};

\node[box, fill=cGreen, draw=cGreenBorder, below=of seq, yshift=-0.1cm, minimum width=11.6cm] (model)
    {Model training (Centralized/Federated)};

\draw[arrow] (parquet) -- (csv);
\draw[arrow] (csv) -- (clean);

\draw[arrow] (trend) -- (norm);
\draw[arrow] (norm) -- (valid);

\draw[arrow] (gov) -- (seq);
\draw[arrow] (seq) -- (pair);

\coordinate (a1) at ($(clean.east)+(0.35,0)$);
\coordinate (aMidY) at ($(clean.south)!0.5!(trend.north)$);
\coordinate (a2) at (a1 |- aMidY);
\coordinate (a3raw) at ($(trend.west)+(-0.35,0)$);
\coordinate (a3) at (a3raw |- a2);
\coordinate (a4) at (a3 |- trend.west);
\draw[arrow] (clean.east) -- (a1) -- (a2) -- (a3) -- (a4) -- (trend.west);

\coordinate (b1) at ($(valid.east)+(0.35,0)$);
\coordinate (bMidY) at ($(valid.south)!0.5!(gov.north)$);
\coordinate (b2) at (b1 |- bMidY);
\coordinate (b3raw) at ($(gov.west)+(-0.35,0)$);
\coordinate (b3) at (b3raw |- b2);
\coordinate (b4) at (b3 |- gov.west);
\draw[arrow] (valid.east) -- (b1) -- (b2) -- (b3) -- (b4) -- (gov.west);

\draw[arrow] (seq.south) -- (model.north);

\end{tikzpicture}}
    \caption{End-to-end governance pipeline.}
    \label{fig:dp}
\end{figure}
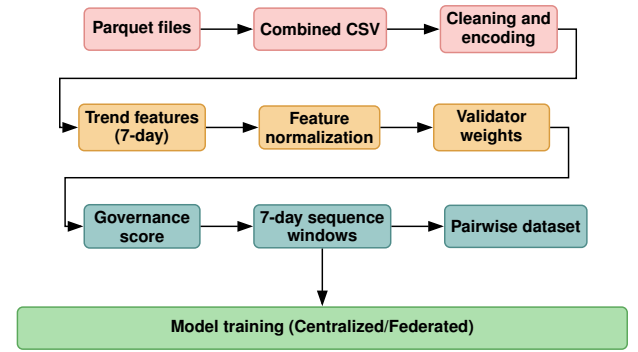

\section{Evaluation}
\label{section:pe}

\subsection{Dataset} 
We collected data from a blockchain platform, specifically the Canton Network, over 184 days (2025-08-24 to 2026-02-23), during which users interact through multiple financial applications. The data is publicly available and was collected from the Super Validator Set of the global synchronizer. The dataset contains transaction logs maintained by host nodes in a decentralized financial structure, where each host stores the activity of the parties it manages. Rather than raw transaction data, the dataset contains aggregated behavioural features derived from financial transaction histories, obtained by processing the raw on-chain records into provider-day aggregates. The data was stored in Parquet file format, where each file contains one day of network activity. All files were merged and exported to CSV format for analysis. The dataset contains 28 features and 965,894 samples, where each row represents a provider on a specific date. Each feature captures a different aspect of provider behaviour, including transaction activity, engagement levels, churn, and network connectivity. Two governance-related binary flags, \textit{is\_featured} and \textit{is\_validator}, indicate the provider's current role in the network.

\subsubsection{Data Preprocessing}
Time-related columns are converted to standard datetime format. Rows with missing \textit{provider} or \textit{record\_date} identifiers are discarded, as these fields are required to construct per-provider temporal sequences. Duplicate records sharing the same \textit{provider} and \textit{record\_date}, arising from overlapping daily snapshots, are removed, retaining only the entry with the most recent \textit{record\_time}. Missing values in the remaining numeric fields are imputed with zero. Numeric columns are cast to the correct data types. Boolean governance flags such as \textit{is\_featured} and \textit{is\_validator} are encoded as 0 and 1.

\subsubsection{Feature Selection}
We selected 13 features that best represent provider stability and governance quality, rather than using all 28. The selected features are: \textit{total\_cc\_volume}, \textit{global\_engagement\_rate}, \textit{interconnectivity}, \textit{reciprocity\_rate}, and \textit{total\_churn\_rate} to capture core transaction and network behaviour; \textit{new\_transfer\_users}, \textit{transfer\_user\_base\_increase}, \textit{unique\_directional\_transfers}, \textit{average\_n\_transfers\_per\_user}, and \textit{new\_beneficiaries} to record user growth dynamics and engagement depth.

Since static features such as \textit{total\_cc\_volume}, \textit{global\_engagement\_rate}, and \textit{transfer\_users} capture only a snapshot in time and do not reflect whether a provider is growing or declining, we additionally define three trend features, computed on the preprocessed data described above, as: 
\begin{equation}
 \text{trend} = \frac{x_t - x_{t-7}}{|x_{t-7}| + 10^{-6}}
\end{equation} 
where $x_t$ is the feature value today and $x_{t-7}$ is the value seven days prior. This yields \textit{volume\_trend}, \textit{engagement\_trend}, and \textit{user\_growth\_trend}. A positive value indicates growth, a negative value indicates decline, and zero indicates no change. For example, if transactions rise from 2,000 to 5,000, the trend is $+1.5$ (150\% growth); if they fall from 9,000 to 5,000, the trend is $-0.44$ (44\% decline). All trend values are clipped to $[-5, +5]$ to prevent extreme outliers. If a provider has fewer than seven days of history, the trend is set to zero. This makes governance evaluation dynamic by giving higher weight to growing providers relative to declining ones.

\subsubsection{Feature Normalization}
The features are normalized using z-score normalization:
\begin{equation}
x' = \frac{x - \mu}{\sigma}
\end{equation}
where $\mu$ and $\sigma$ are the mean and standard deviation computed on the data, normalization statistics are saved to \textit{feature\_mu.csv} and \textit{feature\_sigma.csv} to ensure consistent scaling is applied across all settings.

\subsubsection{Governance Score Computation}
 \label{subsec:gov_score} 

Since the dataset does not contain a ranking column, a governance score is computed as a weighted linear combination of the normalized features and used to generate labels during pairwise dataset construction:
\begin{equation}
\text{gov\_score} = \sum_{i=1}^{n} w_i f_i
\label{eq:gov_score}
\end{equation}
where $f_i$ are the normalized features and $w_i$ are feature weights representing governance importance. Rather than using a single fixed weight vector, weights are generated dynamically to simulate different validator preferences. Starting from a fixed base weight vector, where \textit{reciprocity\_rate} receives the highest base weight of 0.18, reflecting its central importance as a governance signal, small Gaussian noise is added to each weight:
\begin{equation}
w_i = \frac{\max(\bar{w}_i + \epsilon_i,\ 10^{-3})}{\sum_{j=1}^{n} \max(\bar{w}_j + \epsilon_j,\ 10^{-3})}, \quad \epsilon_i \sim \mathcal{N}(0,\ 0.03^2)
\label{eq:validator_weights}
\end{equation}
where $\bar{w}_i$ is the base weight for feature $i$ and $\epsilon_i$ is added noise. The noisy weights are clipped to a minimum of $10^{-3}$ to ensure all features retain a non-negative contribution, then renormalized to sum to 1. This simulates the real-world scenario where different validators assign slightly different importance to governance features, where one validator may prioritize transaction volume while another prioritizes reciprocity. For the federated setting, each client further relabels its local partition of the pairwise pool using a client-specific weight vector, generated by the same noisy-renormalization mechanism but starting from one of five fixed structural base profiles, each emphasizing a single governance dimension, reciprocity, engagement, churn, volume, or interconnectivity:
\begin{equation}
w_i^{(c)} = \frac{\max(p_i^{(c)} + \epsilon_i^{(c)},\ 10^{-3})}{\sum_{j=1}^{n} \max(p_j^{(c)} + \epsilon_j^{(c)},\ 10^{-3})}, \quad \epsilon_i^{(c)} \sim \mathcal{N}(0,\ 0.02^2)
\label{eq:client_weights}
\end{equation}
where $p^{(c)}$ is the fixed base profile for client $c$, with the corresponding governance dimension assigned a base weight of 0.35-0.40 and the remaining features assigned small residual weights. This is the mechanism referenced in Section~\ref{section:pm}: since each client starts from a different $p^{(c)}$, the same provider pair can be assigned opposite labels by different clients.

The governance score is used only to generate labels during pairwise dataset construction and is never used as a direct model input.

\subsubsection{Pairwise Dataset Construction}

Generating all possible provider pairs is computationally infeasible given 965,894 samples. We therefore adopt a sampling-based pairwise approach. For each \textit{record\_date} (governance round), providers active on that day are sampled in pairs without restriction on featured status, allowing non-featured-versus-featured, featured-versus-featured, and non-featured-versus-non-featured comparisons to all occur. The number of sampled pairs per day is capped at $\min(5N_t,\ 5000)$, where $N_t$ is the number of providers active on day $t$, to bound memory usage while scaling pair count with daily provider activity.

For each sampled pair $(i, j)$, governance scores are compared, and the pair is discarded if $|\text{gov\_score}_i - \text{gov\_score}_j| < 10^{-6}$ to avoid near-tied comparisons that provide no meaningful ranking signal. Otherwise, a single row is written with label $1$ if $\text{gov\_score}_i > \text{gov\_score}_j$, and label $0$ otherwise, with provider $i$'s features placed in columns suffixed \textit{\_1} and provider $j$'s in columns suffixed \textit{\_2}.

Each side of a pair stores the full 7-day rolling sequence described in Section~\ref{subsec:local_training} rather than a single-day snapshot, yielding $13 \times 7 = 91$ feature columns per side ($182$ feature columns total) plus the label column. This sampling procedure, applied across all governance rounds in the dataset, produces a final pairwise dataset of 9,657,042 record pairs.

\subsection{Experimental Setup}
\label{subsec:Setup}
All experiments ran on a Compute Canada \textit{Nibi} node\footnote{\url{https://docs.alliancecan.ca/wiki/Nibi}} (NVIDIA H100 80GB) using the Flower framework~\cite{beutel2007flower}. The pairwise dataset is split 80:20 for training/test, with a separate date-level 80:20 holdout for ranking evaluation to prevent temporal leakage. Training uses batch size 2048, Adam ($\eta=0.001$), and \textit{BCEWithLogitsLoss}. Each provider's input is a 7-day rolling window of 13 governance features, shape $(N,7,13)$, used for both training and evaluation; NDCG@5/10/15 and Kendall $\tau$ are computed each epoch against the governance score ground truth. Centralized training runs 100 epochs; federated training simulates 5 clients over 100 rounds, with non-IID heterogeneity introduced via a Dirichlet distribution ($\alpha=0.5$) and each client relabeling its partition with a client-specific structural weight profile.
\subsection{Model Architecture}
\label{sec:model}
\begin{figure}[t]
    \centering
    \includegraphics[width=\linewidth]{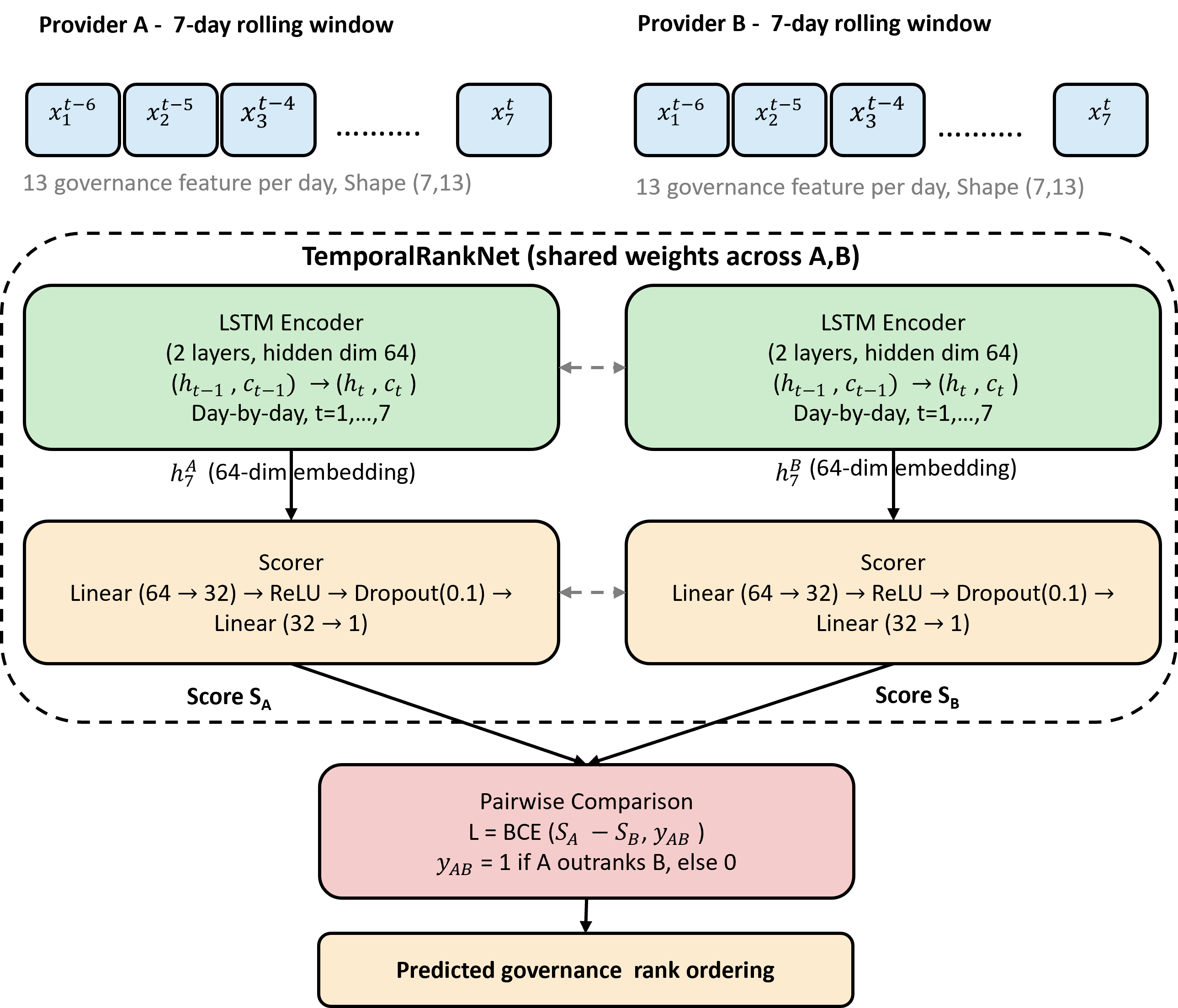}
    \caption{TemporalRankNet architecture: shared LSTM encoder and scorer map each provider's 7-day feature window to a score, trained via pairwise BCE loss.}
    \label{fig:temporalranknet_arch}
\end{figure}

Both architectures follow the same pairwise formulation~\cite{burges2005learning}: each provider's 7-day, 13-feature sequence is encoded into a single embedding, mapped to a scalar score, and compared between two same-day providers. \textbf{TemporalRankNet.} A two-layer LSTM (hidden dim 64) processes the sequence day by day; the final hidden state $h_7$ is mapped to a score via Linear($64{\to}32$), ReLU, Dropout($0.1$), Linear($32{\to}1$) (Fig.~\ref{fig:temporalranknet_arch}). \textbf{AttentionRankNet.} A 4-head self-attention encoder with sinusoidal positional encoding processes the 7-day window; the mean-pooled output is mapped to a score via a 3-layer MLP head. 
Both models optimize Eq.~\ref{eq:pairwise_loss} over 9,657,042 training pairs, where $y_{AB}=1$ if provider $A$ has higher governance quality than $B$, and 0 otherwise.

\subsection{Performance Metrics}
\label{sec:performance_metrics}
We evaluate ranking quality using Normalized Discounted Cumulative Gain (NDCG) and the Kendall rank correlation coefficient $\tau$. The model is optimized using the pairwise ranking loss defined in Eq.~\ref{eq:pairwise_loss}, followed by evaluation using the two ranking metrics below.

\subsubsection{Normalized Discounted Cumulative Gain}
NDCG evaluates both the relevance and position of items in the ranked list~\cite{jarvelin2002cumulated}, penalizing ranking errors at the top of the list more heavily, well-suited to governance decisions where top-ranked providers receive featured status and greater network visibility. NDCG values range from 0 to 1, with higher values indicating better ranking quality. We report NDCG@5, NDCG@10, and NDCG@15.

\subsubsection{Kendall $\tau$}
Kendall $\tau$~\cite{kendall1938new} measures the rank correlation between predicted and ground-truth orderings across all provider pairs:
\begin{equation}
\tau = \frac{C - D}{\binom{n}{2}}
\label{eq:kendall_tau}
\end{equation}
where $C$ and $D$ are the numbers of concordant and discordant pairs among all $\binom{n}{2}$ provider pairs, and $\tau \in [-1, 1]$. While NDCG@K emphasizes top-ranking precision, Kendall $\tau$ captures overall pairwise ranking consistency across the full provider list, offering a complementary view of global ranking quality.

\subsection{Baselines}
We compare two training regimes for both architectures (Section~\ref{sec:model}). \textbf{Centralized}: trained on the full dataset at a single server, with direct access to all provider data. \textbf{Federated}: each client trains locally and shares only model updates without exposing raw transaction data. Crossing both regimes with both architectures yields four settings: Centralized LSTM-RankNet (C-L), Federated LSTM-RankNet (F-L), Centralized Attention-RankNet (C-A), and Federated Attention-RankNet (F-A).

\begin{figure*}[htbp]
    \centering
    \setkeys{Gin}{width=.49\linewidth}
    \begin{minipage}[t]{1\columnwidth}
        \subfloat[Centralized NDCG]{\includegraphics{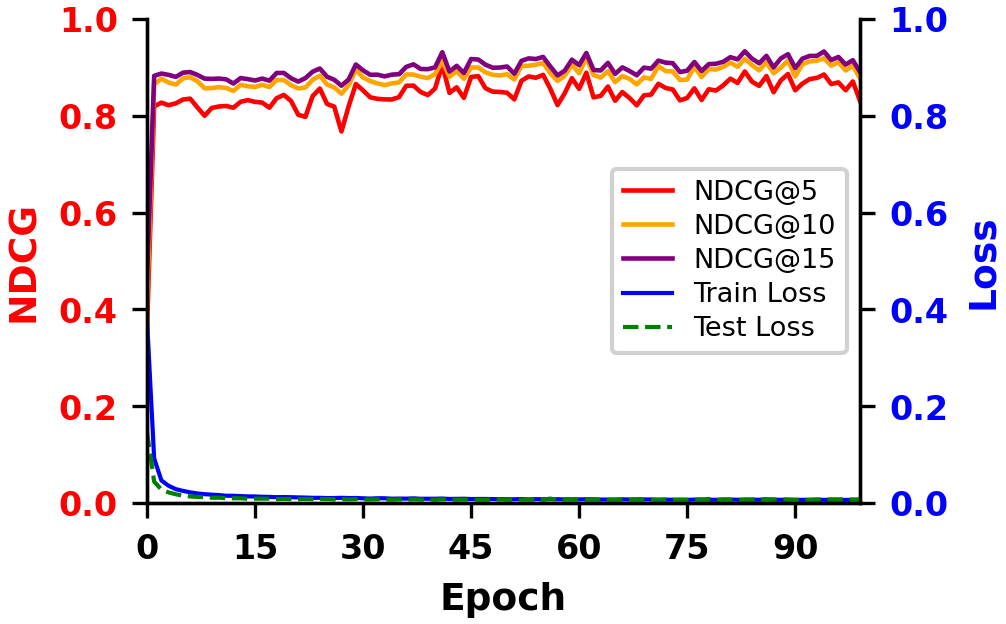}}\hfill
        \subfloat[Centralized Kendall]{\includegraphics{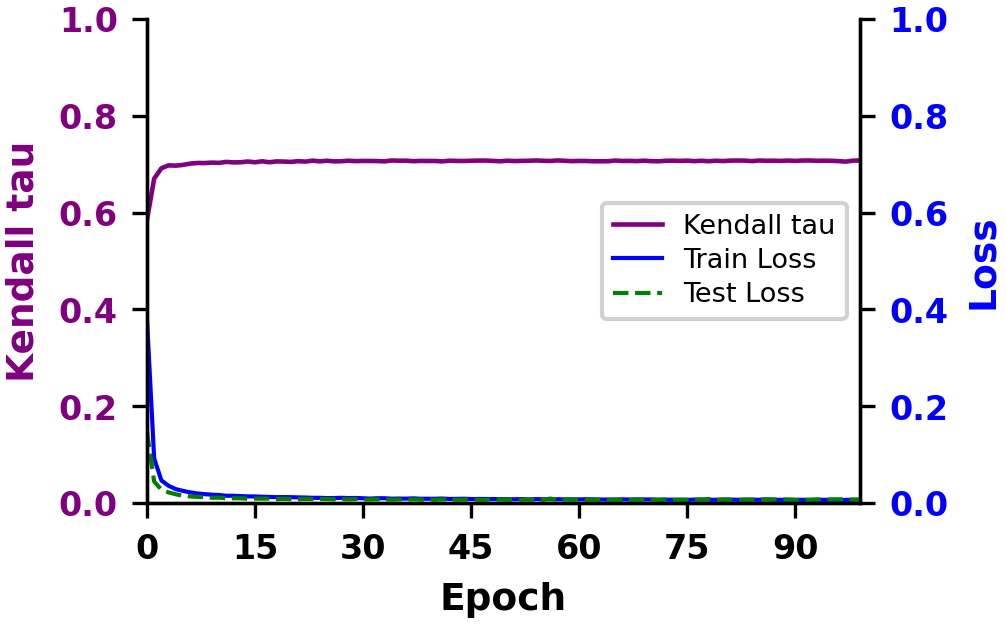}}
    \end{minipage}\hfill
    \begin{minipage}[t]{1\columnwidth}
        \subfloat[Decentralized NDCG]{\includegraphics{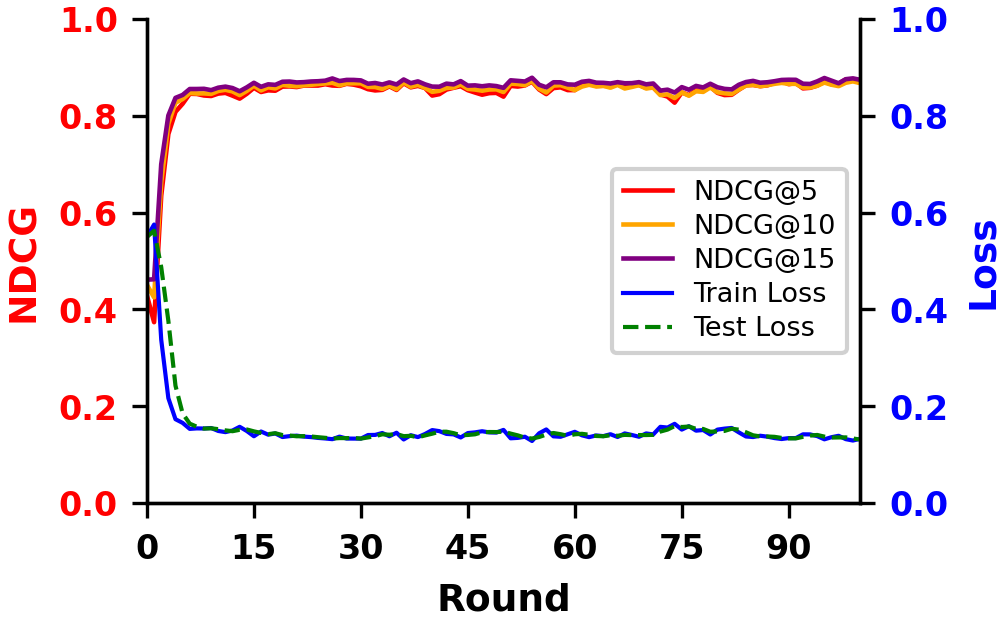}}\hfill
        \subfloat[Decentralized Kendall]{\includegraphics{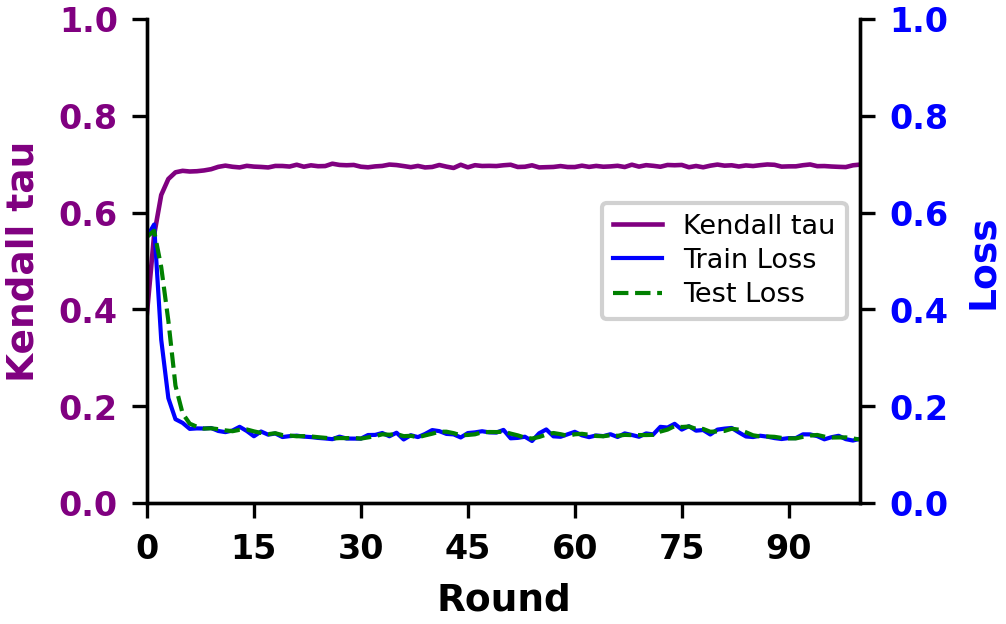}}
    \end{minipage}
    \caption{LSTM-RankNet}
    \label{fig:LSTM-RankNet}
\end{figure*}

\begin{figure*}[htbp]
    \centering
    \setkeys{Gin}{width=.49\linewidth}
    \begin{minipage}[t]{1\columnwidth}
        \subfloat[Centralized NDCG]{\includegraphics{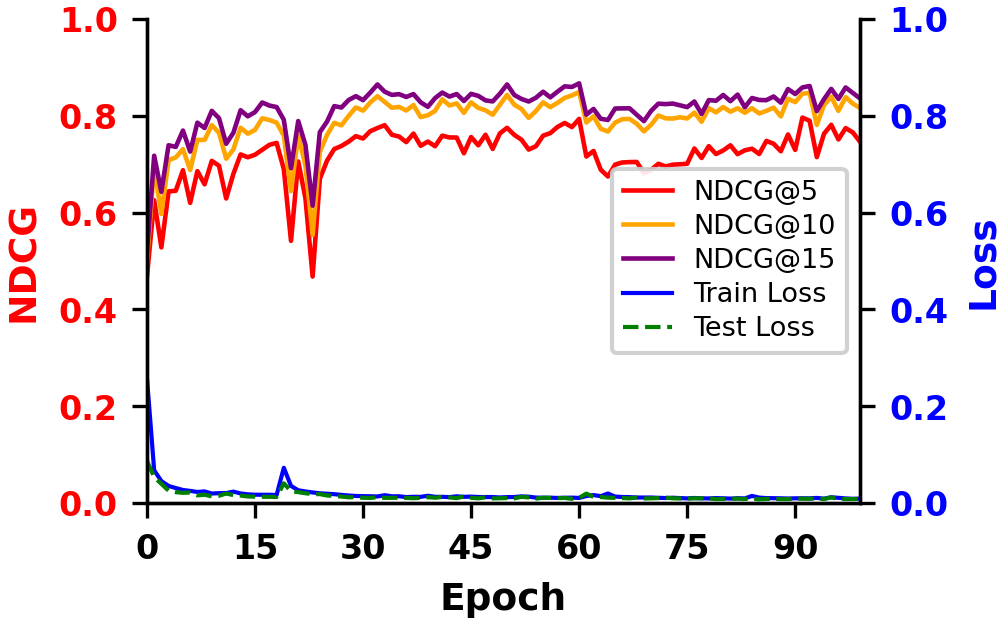}}\hfill
        \subfloat[Centralized Kendall]{\includegraphics{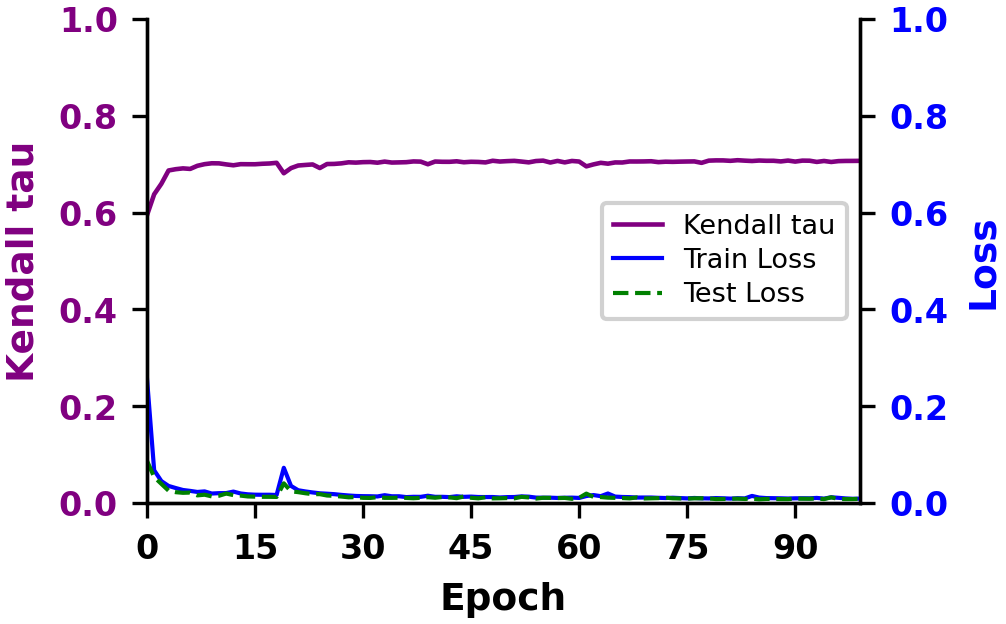}}
    \end{minipage}\hfill
    \begin{minipage}[t]{1\columnwidth}
        \subfloat[Decentralized NDCG]{\includegraphics{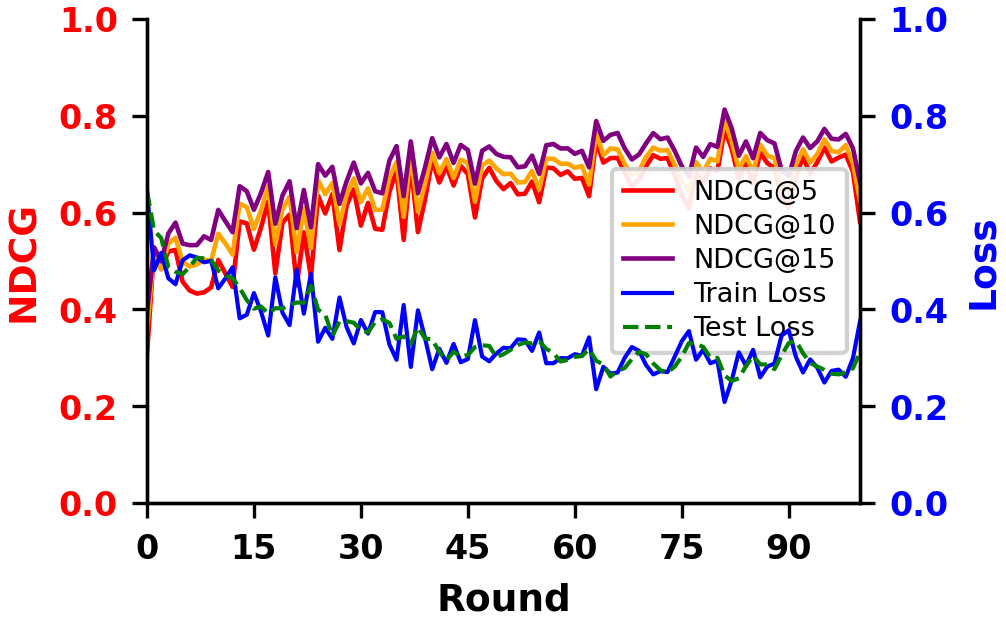}}\hfill
        \subfloat[Decentralized Kendall]{\includegraphics{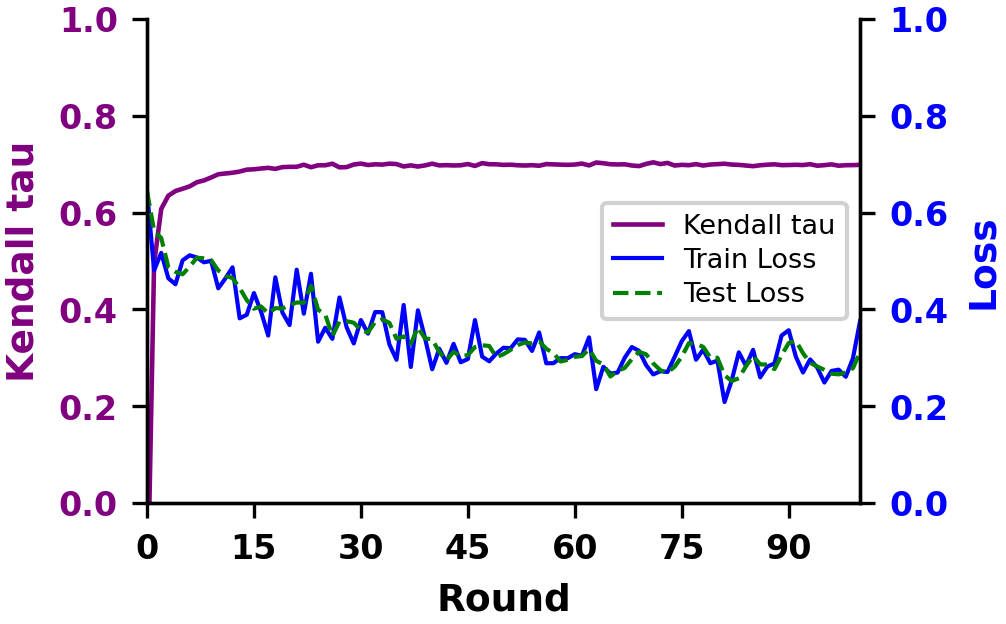}}
    \end{minipage}
    \caption{Attention-RankNet}
    \label{fig:Attention-RankNet}
\end{figure*}

\subsection{Results}
\label{sec:results}
Fig.~\ref{fig:LSTM-RankNet} (a),(b) report centralized LSTM-RankNet performance. NDCG increases in a steady high range for the duration of training to 0.82-0.88 in a single epoch after getting nearly random initial values. The final values for NDCG@k are (0.87, 0.91, 0.92) with loss converging to (0.006-0.007) by epoch~100 and no train/test gap. Kendall~$\tau$ converges within 5-6 epochs at 0.70, since NDCG@k is sensitive to reordering near the cutoff while $\tau$ aggregates over all pairs, smoothing local noise.

Fig.~\ref{fig:LSTM-RankNet}(c),(d) report federated performance. Round~1 shows a temporary drop before recovering (NDCG@5: 0.43$\to$0.37, $\tau$: 0.39$\to$0.55): each client weights the 13 governance features differently (Section~\ref{section:pm}), so the same provider pair can receive opposite labels across clients, and the first aggregation must reconcile these conflicting local models before a stable global ranking emerges. NDCG@k converges by round~6 to (0.86,0.87, 0.88). $\tau$ converges by round~6 at 0.70, comparable to the centralized.

Fig.~\ref{fig:Attention-RankNet}(a),(b) report centralized Attention-RankNet performance. NDCG rises to 0.63-0.72 within two epochs but continuously fluctuating and less stable than LSTM-RankNet. NDCG plateaus lower and wider than LSTM-RankNet with final values (0.75, 0.82, 0.84) below LSTM-RankNet's peak ($\approx$0.92 NDCG@15). However, $\tau$ is unaffected and converging to 0.706, similar to LSTM-RankNet.

Fig.~\ref{fig:Attention-RankNet}(c),(d) report federated Attention-RankNet performance. NDCG@5 starts near-random and shows continuous fluctuations. $\tau$ converges similar to LSTM-RankNet. NDCG@k continues oscillating without a clear trend ending at round~100 with (0.72, 0.74, 0.77) and $\tau$=0.70, both below LSTM-RankNet's federated plateau.

Federated training differs from centralized in one key way: each client scores providers using its own feature weights, while centralized uses one shared weight vector (Section~\ref{section:pm}). This applies equally to both architectures, so it cannot explain why Attention-RankNet converges roughly five times slower than LSTM-RankNet under federation. The likely reason is architectural: LSTM naturally emphasizes recent days, which suits the 7-day trend features, while Attention averages all days equally, diluting that signal, especially when clients disagree on which features matter. Attention also has more parameters than a 7-day sequence really needs, which likely explains its training instability under centralized learning too. These results motivate using LSTM-RankNet for the governance decision pipeline (Section~\ref{subsec:Governance_decision_making}).

Table~\ref{tab:anchor_summary} summarizes the baseline anchors and final candidate counts for centralized and federated LSTM-RankNet. Both compute their anchors from their own baseline period (Section~\ref{subsec:Governance_decision_making}), so the two settings use independently data-driven thresholds rather than a shared cutoff. Centralized has lower promotion and minimum ratio anchors, which reflect scores spanning the full corpus and settling into a narrower baseline distribution, whereas federated has stricter anchors, including the promotion anchor saturating at its upper clip bound, which reflect clients' conflicting weight vectors (Section~\ref{section:pm}) producing a baseline score distribution that is higher and tighter, which the reliability filter (Algorithm~\ref{alg:governance}, line~\ref{line:reliability}) turns into a stricter bar.

In the centralized setting (Table~\ref{tab:anchor_summary}), all qualifying promotion candidates showed fully consistent signals over 2+ days, suggesting the model reliably learns well-governed non-featured applications under the looser anchor. Demotion candidates split roughly two-thirds consistently poor (low transaction quality, high churn, low reciprocity) to one-third borderline, appearing only once.

In the federated setting (Table~\ref{tab:anchor_summary}), far fewer providers qualify for promotion than centralized, but at a higher average ratio, consistent with the stricter federated anchor: fewer providers clear the bar, but those that do outscore a larger share of active featured competitors. The demotion anchor and top-$K$ bounds (Algorithm~\ref{alg:governance}, lines~\ref{line:demotion_anchor} and \ref{line:effective_topk}) are identical across both regimes, since they depend only on the elbow position in the sorted featured-app scores and the network-size ratio, not the validator weighting scheme, reflected in both settings flagging the same number of demotion rows. Despite the shared anchor, federated demotion shows marginally higher consistency than centralized, suggesting federated aggregation produces slightly more temporally stable demotion signals, consistent with the tighter Kendall~$\tau$ band observed for federated training.


\begin{table}[t]
\centering
\caption{Baseline anchors, final candidate counts, Promotion and demotion candidate statistics for Centralized LSTM-RankNet and Federated LSTM-RankNet.}
\label{tab:anchor_summary}
\footnotesize
\begin{tabular}{lcc}
\toprule
\textbf{Metric} & \textbf{Centralized} & \textbf{Federated} \\
\midrule
\rowcolor{green!15}
Promotion anchor $\tau_{\text{promo}}$ & 0.445 & 0.600 (ceiling) \\
\rowcolor{green!15}
Min ratio anchor $\rho_{\text{min}}$ & 0.546 & 0.700 \\
\rowcolor{green!15}
Baseline qualifying apps & 1,301 & 1,212 \\
\rowcolor{red!15}
Demotion anchor $K_{\text{base}}$ & 150 (ceiling) & 150 (ceiling) \\
\midrule
\rowcolor{green!15}
Reliable promotion candidates & 15,587 & 2,257 \\
\rowcolor{green!15}
Unique promotion providers & 4,009 & 648 \\
\rowcolor{green!15}
Mean promotion ratio & 0.645 & 0.772 \\
\rowcolor{green!15}
Consistent (2+ days) & 4,009 (100\%) & 648 (100\%) \\
\midrule
\rowcolor{red!15}
Demotion candidate rows & 183 & 183 \\
\rowcolor{red!15}
Unique demotion providers & 35 & 39 \\
\rowcolor{red!15}
Consistent (2+ days) & 24 (69\%) & 28 (72\%) \\
\rowcolor{red!15}
Borderline (1 day) & 11 (31\%) & 11 (28\%) \\
\bottomrule
\end{tabular}
\end{table}

\begin{figure}[htbp]
    \centering
    \begin{subfigure}[b]{0.23\textwidth}
        \includegraphics[width=\textwidth]{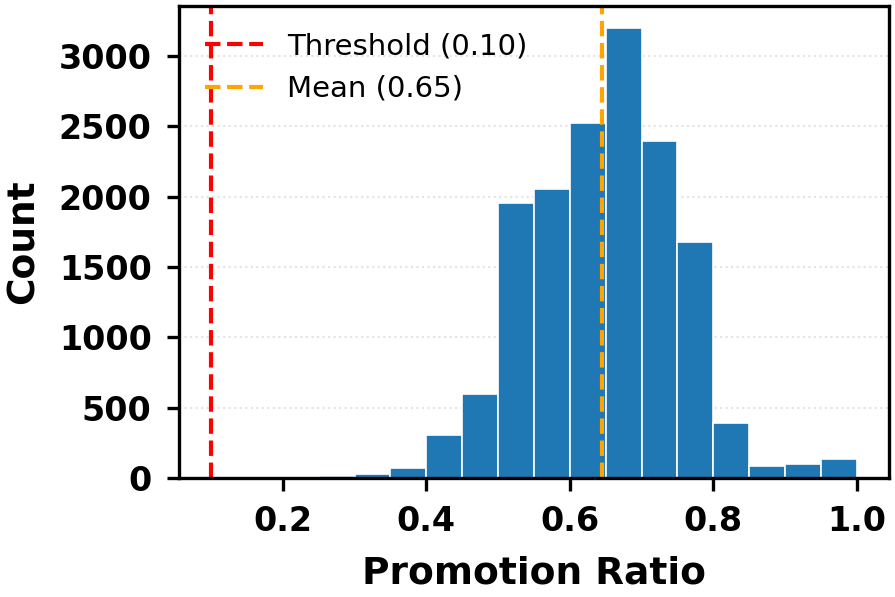}
        \caption{Centralized}
        \label{fig:centralized_promotion_ratio_distribution}
    \end{subfigure}
    \hfill
    \begin{subfigure}[b]{0.23\textwidth}
        \includegraphics[width=\textwidth]{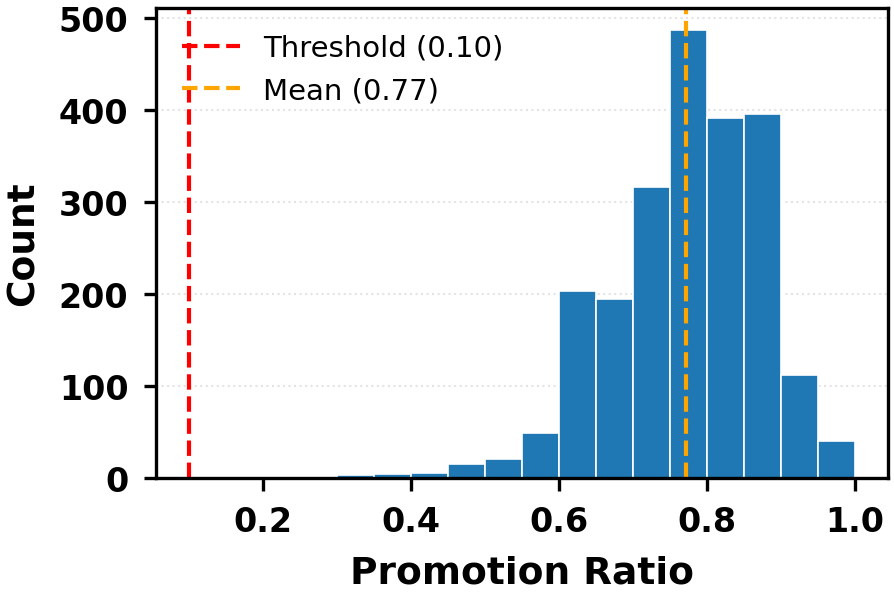}
        \caption{Decentralized}
        \label{fig:decentralized_promotion_ratio_distribution}
    \end{subfigure}
    \vspace{-0.3em}
    \caption{LSTM-RankNet promotion ratio distribution comparison}
    \label{fig:promotion_ratio_comparison}
\end{figure}

Fig.~\ref{fig:centralized_promotion_ratio_distribution} shows the promotion ratio distribution for centralized LSTM-RankNet. The dashed red line marks $\tau_t=0.10$, the lower bound of the promotion threshold's clipped range (Algorithm~\ref{alg:governance}, line~\ref{line:promo_threshold}); the actual daily threshold varies within this range depending on network health, so this line is the most permissive value $\tau_t$ can take, not a fixed cutoff. The distribution concentrates between 0.3 and 1.0, peaking near 0.65-0.7, with the dashed orange line marking the mean (Table~\ref{tab:anchor_summary}): a promoted provider outscores active featured apps well above both this floor and the minimum ratio anchor (Table~\ref{tab:anchor_summary}) on average. Few providers fall in 0.3-0.4, indicating the reliability filter successfully separates sustained quality differentials from transient threshold crossings.

Fig.~\ref{fig:decentralized_promotion_ratio_distribution} shows the federated distribution, right-shifted and narrower, peaking near 0.75-0.80, with far fewer total candidates than centralized (Table~\ref{tab:anchor_summary}). Both differences trace to federated's stricter minimum ratio anchor, itself a consequence of clients' conflicting structural weight vectors (Section~\ref{section:pm}) producing a baseline score distribution concentrated higher and tighter than the centralized full-corpus baseline. Since the reliability filter rejects any candidate below this anchor, a stricter anchor mechanically raises the surviving population's mean ratio while shrinking its size, exactly the pattern observed. Neither distribution shows providers clustering near the 0.10 floor, confirming the filter is effective regardless of the anchor value a baseline period yields.

\begin{figure}[htbp]
    \centering
    \begin{subfigure}[b]{0.23\textwidth}
        \includegraphics[width=\textwidth]{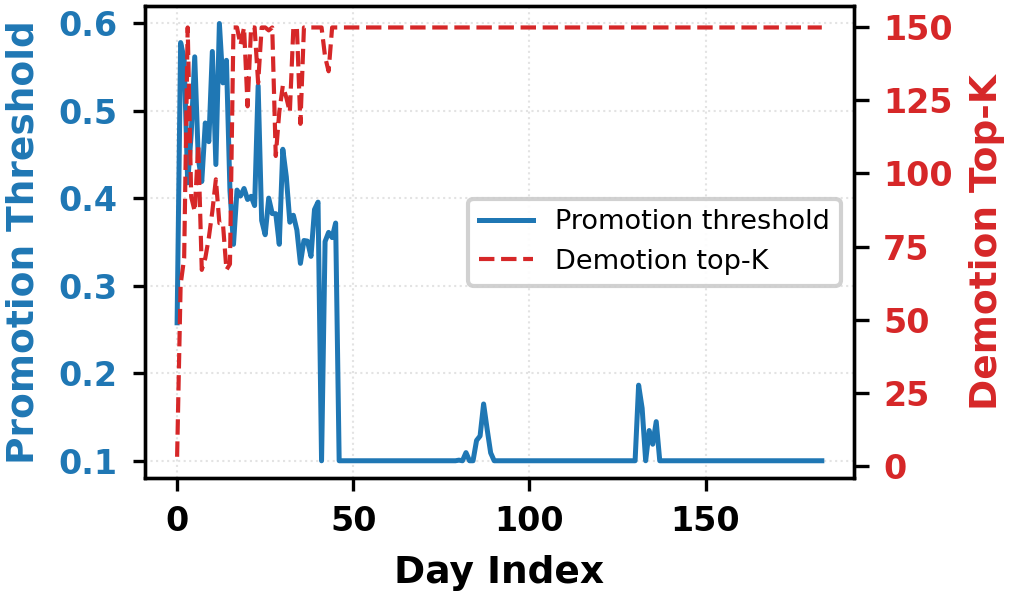}
        \caption{Centralized}
        \label{fig:centralized_dynamic_thresholds}
    \end{subfigure}
    \hfill
    \begin{subfigure}[b]{0.23\textwidth}
        \includegraphics[width=\textwidth]{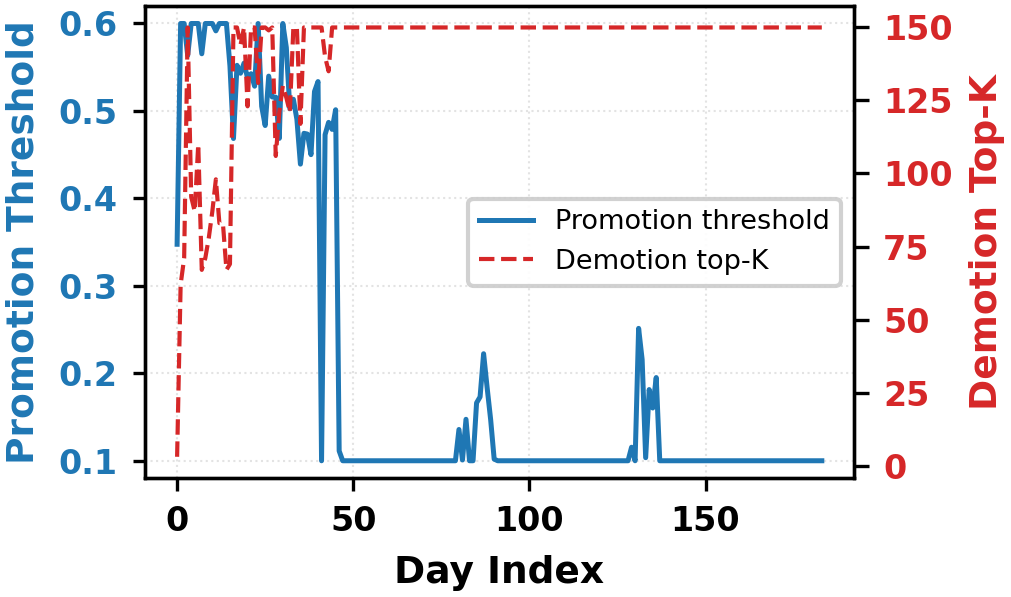}
        \caption{Decentralized}
        \label{fig:decentralized_dynamic_thresholds}
    \end{subfigure}
    \vspace{-0.3em}
    \caption{LSTM-RankNet Centralized and Decentralized thresholds comparison}
    \label{fig:dynamic_thresholds}
\end{figure}

Fig.~\ref{fig:dynamic_thresholds}(a) and (b) show the daily promotion threshold ($\tau_t$) and demotion cutoff ($K_t$) for the centralized and federated models. Both use the same health and size ratios from \textit{gov\_score}; only the anchor values differ because they are computed from each model's predictions.

The demotion curves are identical because both models reach the maximum $K_{\text{base}}=150$, so $K_t=150\times\lambda_t^{\text{size}}$. The promotion thresholds differ because the federated anchor is at the maximum value, while the centralized anchor is lower.

After day 40, both thresholds stay at the minimum ($\tau_t=0.10$) as network health declines and network size grows rapidly. Small peaks around days 87 and 131 appear in both models due to improved health values, but the federated model reaches slightly higher peaks because its larger anchor makes it more sensitive near the minimum threshold.

\section{Conclusion} 
\label{sec:conclusion} 
We presented \name, a privacy-preserving federated pairwise ranking framework that replaces static, manually-crafted epoch scoring functions in decentralized financial networks. By training TemporalRankNet, combining LSTM temporal encoding with RankNet pairwise ranking across 13 governance features, over 9.6 million pairwise comparisons distributed across five non-IID clients, \name\ learns network-wide behavioural patterns without exposing private transaction data. We additionally evaluated an attention-based architecture, AttentionRankNet, which achieved comparable Kendall~$\tau$ but consistently lower and less stable NDCG, motivating LSTM-RankNet as the architecture used for governance decisions. A dynamic governance pipeline adaptively adjusts promotion and demotion thresholds in response to network health and size, yielding 648 unique promotion candidates and 39 unique demotion candidates under federation, with 100\% promotion consistency and 72\% demotion consistency over 2+ consecutive days. Experimental results on a real-world blockchain dataset show that \name\ achieves ranking performance modestly below the centralized baseline (NDCG@15: $\approx$0.87 vs.\ $\approx$0.92) while reaching comparable ranking stability (Kendall~$\tau$: $\approx$0.70 for both regimes), demonstrating that federated learning preserves institutional privacy without sacrificing the consistency of governance signals. Future work will explore integrating tokenomic incentive mechanisms to couple governance decisions with on-chain rewards, adopting personalised federation strategies to better accommodate non-IID client heterogeneity, and incorporating formal differential privacy guarantees for deployment in regulated financial environments.

\section*{Acknowledgment}
This research is supported by T-RIZE GROUP, Montréal, Canada, and Mitacs Grant IT44869.

\bibliographystyle{IEEEtran}
\bibliography{IEEEabrv,biblio_traps_dynamics}

\end{document}